\pdfoutput=1

\documentclass[11pt]{article}

\usepackage[preprint]{acl}

\usepackage{times}
\usepackage{latexsym}

\usepackage[T1]{fontenc}

\usepackage[utf8]{inputenc}

\usepackage{microtype}

\usepackage{inconsolata}

\usepackage{graphicx}

\usepackage{amsmath,amsfonts,graphicx,amssymb,bm,amsthm}
\usepackage{bbding}
\usepackage{booktabs}
\usepackage{tcolorbox}

\definecolor{green}{RGB}{5,146,18} 
\definecolor{orange}{RGB}{240,135,5}
\definecolor{blue}{RGB}{0,204,255}

\title{A Probabilistic Inference Scaling Theory for LLM Self-Correction}

\author{Zhe Yang$^{1}$, Yichang Zhang$^{2}$, Yudong Wang$^{1}$, Ziyao Xu$^{1}$, Junyang Lin$^{2}$, Zhifang Sui$^{1}$\thanks{Corresponding author} \\
$^1$State Key Laboratory of Multimedia Information Processing, \\
School of Computer Science, Peking University \\ 
$^2$Alibaba Group \\
\texttt{\{yz\_young, szf\}@pku.edu.cn}\\
}

\begin{document}
\maketitle
\begin{abstract}

Large Language Models (LLMs) have demonstrated the capability to refine their generated answers through self-correction, enabling continuous performance improvement over multiple rounds. However, the mechanisms underlying how and why accuracy evolves during this iterative process remain unexplored. 
To fill this gap, we propose a probabilistic theory to model the dynamics of accuracy change and explain the performance improvements observed in multi-round self-correction. 
Through mathematical derivation, we establish that the accuracy after the $t^{th}$ round of self-correction is given by: $Acc_t = Upp - \alpha^t(Upp - Acc_0),$
where $Acc_0$ denotes the initial accuracy, $Upp$ represents the upper bound of accuracy convergence, and $\alpha$ determines the rate of convergence. 
Based on our theory, these parameters can be calculated and the predicted accuracy curve then can be obtained through only a single round of self-correction. 
Extensive experiments across diverse models and datasets demonstrate that our theoretical predictions align closely with empirical accuracy curves, validating the effectiveness of the theory. 
Our work provides a theoretical foundation for understanding LLM self-correction, thus paving the way for further explorations.

\end{abstract}

\section{Introduction}

With the depletion of pre-training corpora, the training scaling law \citep{kaplan2020scaling} reaches the saturation point, and an alternative way to further improve performance is introducing more computational cost at test time, also known as inference scaling \citep{snell2025scaling,hoffmann2022training}. 
\citet{brown2024largelanguagemonkeysscaling} repeatedly sample multiple answers and select the optimal one with best-of-\textit{n} \citep{li-etal-2023-making} or majority voting \citep{wang2023selfconsistency} strategy, and the curve of how accuracy changes in this process as inference costs increase is also experimentally recorded \citep{wu2024scaling}.
Another approach to inference scaling is self-correction \citep{kamoi-etal-2024-llms,pan-etal-2024-automatically}, where LLMs can refine their answers based on intrinsic \citep{madaan2024self} or external \citep{jiang2023selfevolve} feedback. 
\citet{xi-etal-2023-self,liu2024intrinsicselfcorrectioncapabilityllms} have empirically observed that model performance continuously improves and eventually converges during multi-round self-correction, but the underlying reasons and mechanisms remain poorly understood.
To narrow this gap, we propose a probabilistic theory to model how accuracy evolves and explain why performance improves in multi-round self-correction.

In \S \ref{sec:theory}, we mathematically derive our theory from a probabilistic perspective.
\citet{yang2024decomposition} decompose self-correction capabilities of LLMs into confidence capability and critique capability, introducing two metrics named Confidence Level ($CL$) and Critique Score ($CS$) to measure them, respectively.
Based on their decomposition, we further discover a recursive relationship between the accuracy of successive rounds of self-correction: $Acc_{t} = (CL-CS)Acc_{t-1} + CS$, where $Acc_{t}$ and $Acc_{t-1}$ denote the accuracy after the $t^{th}$ and ${t-1}^{th}$ round of self-correction, respectively.
From this recursive relationship, we further find $Acc_t = Upp - \alpha^t( Upp - Acc_0)$, where $Upp = \frac{CS}{1-CL+CS}, \alpha=CL-CS$, and $Acc_0$ is the initial accuracy. 
This equation serves as the core part of our theory by describing how accuracy evolves in multi-round self-correction. Further, we derive several corollaries about converged accuracy and convergence rate.

To directly verify the theory, we compare the empirical accuracy curve with the theoretical curve given by our theory, and extensive experiments in \S \ref{sec:experiment} demonstrate that the theoretical curve fits the empirical curve well across various models and datasets. Besides, we also give empirical verification of 3 corollaries as further support for our theory in Appendix \ref{sec:corollaries}.

Our contributions can be summarized as follows:
\begin{enumerate}

\item We propose a probabilistic theory to model how accuracy evolves in multi-round self-correction, along with 3 corollaries.

\item To validate our theory, we conduct extensive experiments and find that our theoretical curve fits empirical curve well.

\item Our theory provides a probabilistic perspective to better understand self-correction.

\end{enumerate}

\section{Theory}
\label{sec:theory}
In this section, we introduce an inference scaling theory to model and explain how accuracy changes in multi-round self-correction. First, we formally define the multi-round self-correction process and provide mathematical notations in \S \ref{subsec:notations}. 
Then we discuss a simple scenario where the test set consists of only one datum (\S \ref{subsec:question_level}), 
and further extend our analysis to the general case where the test set contains $n$ questions (\S \ref{subsec:dataset_level}). 
According to our theory, the accuracy after $t$ rounds of self-correction is given by $Acc_t = Upp - \alpha^t( Upp - Acc_0)$ and finally converges to $Upp$. 
Besides, we also give three corollaries for our theory in \S \ref{subsec:corollaries}.
\subsection{Problem Formulation and Notations}
\label{subsec:notations}
Initially, we have a set comprising of $n$ questions denoted as $Q = \{ q_1, q_2, ..., q_n \}$, and we utilize multi-round self-correction to boost model performance.
For any given question $q_i$, we first directly query the model and generate an answer $a_{i,0}$. Then we utilize an appropriate prompt to encourage the model to self-correct $a_{i,0}$ and get a refined answer $a_{i,1}$ and subsequently self-correct $a_{i,1}$ to get $a_{i,2}$, and so on. 
This process is conducted iteratively, yielding a sequence of answers $a_{i,0}, a_{i,1}, ..., a_{i,k}$ after $k$ rounds of self-correction.
For the answer $a_{i,t}$ from the $t^{th}$ self-correction, we denote the probability that the model generates a correct answer through a single temperature-based sampling as $P(a_{i,t})$. 
The initial accuracy is defined as $Acc_0=\frac{\sum_{i=1}^{n} P(a_{i,0})}{n}$, and the accuracy after the $t^{th}$ self-correction round is defined as $Acc_t=\frac{\sum_{i=1}^{n} P(a_{i,t})}{n}$. 
For clarity, all notations and their corresponding definitions are summarized in Appendix \ref{app:notations}.

\subsection{Question-Level Theory}
\label{subsec:question_level}

We first discuss how the probability of generating a correct answer for a single question $q_i$ evolves as the number of self-correction rounds increases.
For answer $a_{i,t}$ generated in the $t^{th}$ self-correction, the answer before self-correction $a_{i,{t-1}}$ may be either correct or wrong, so by the Law of Total Probability we have:
\begin{equation}
\begin{aligned}
\label{equ:total_probability}
    P(a_{i,{t}}) =& P(a_{i,t-1})P(a_{i,{t}}|a_{i,{t-1}}) \\
    &+ [1-P(a_{i,t-1})]P(a_{i,{t}}|\neg a_{i,{t-1}}),
\end{aligned}
\end{equation}

where $P(a_{i,{t}}|a_{i,{t-1}})$ and $P(a_{i,{t}}|\neg a_{i,{t-1}})$ denote the conditional probabilities that $a_{i,t}$ is correct given that $a_{i,{t-1}}$ is correct or incorrect, respectively. 
During the $t^{th}$ self-correction round, only $a_{i,{t-1}}$ is fed into the model, rather than the whole sequence $a_{i,0},..., a_{i,{t-1}}$.
Consequently, these two probabilities depend solely on the question index $i$ and are independent of the current self-correction round $t$.
We denote these two probabilities as $P_i^{con}$ and $P_i^{cri}$ respectively, which represent the probability of generating a correct answer after self-correction, given the answer before self-correction is correct/wrong. 
For any $t \in N+$, we have $P(a_{i,t}|a_{i,t-1})=P_i^{con}$ and $P(a_{i,t}|\neg a_{i,t-1})=P_i^{cri}$, which we substitute into Equation \ref{equ:total_probability} to obtain:
\begin{equation}
\begin{aligned}
\label{equ:recursion_equation}
P(a_{i,{t}}) &= P(a_{i,t-1})P_i^{con} + [1-P(a_{i,t-1})]P_i^{cri} \\
 &= (P_i^{con}-P_i^{cri})P(a_{i,t-1}) + P_i^{cri} 
\end{aligned}
\end{equation}

It can be further derived (details are shown in Appendix \ref{app:math_derivation}):

\begin{equation}
\begin{aligned}
P(a_{i,{t}}) &= P_i^{upp} - \alpha_i^t( P_i^{upp} -P(a_{i,0}))
\label{equ:final_question_level}
\end{aligned}
\end{equation}

where $P_i^{upp} = \frac{P_i^{cri}}{1-P_i^{con}+P_i^{cri}}$ is the upper bound accuracy converges to, and $\alpha_i = P_i^{con}-P_i^{cri}$ determines the convergence rate.

\subsection{Dataset-Level Theory}
\label{subsec:dataset_level}

Further we try to extend the question-level theory in \S \ref{subsec:question_level} to dataset-level.
\citet{yang2024decomposition} decompose the self-correction capability of a model into two components: confidence (the ability to maintain confidence in the correct answer) and critique (the ability to correct wrong answers), and propose two probabilistic metrics to measure these capabilities quantitatively, which we adopt directly:

\textbullet~ The \textbf{C}onfidence \textbf{L}evel (\textit{CL}) measures the model confidence, defined as the probability that the model retains the correct answer after self-correction:
\begin{equation}
\begin{aligned}
\label{equ:CL}
CL_t &= E[P(a_{\_,t+1}|a_{\_,t})] \\
&= \frac{\sum_{i=1}^{n}P(a_{i,{t}})P(a_{i,{t+1}}|a_{i,t})} {\sum_{i=1}^{n}P(a_{i,t})},
\end{aligned}
\end{equation}

\textbullet~ The \textbf{C}ritique \textbf{S}core (\textit{CS}) measures the capability to critique and reflect, defined as the probability that the model corrects a wrong answer to a right one after self-correction:
\begin{equation}
\begin{aligned}
\label{equ:CS}
CS_t &= E[P(a_{\_,t+1}|\neg a_{\_,t})] \\
&= \frac{\sum_{i=1}^{n}[1-P(a_{i,{t}})]P(a_{i,{t+1}}|\neg a_{i,t})} {\sum_{i=1}^{n}[1-P(a_{i,t})]},
\end{aligned}
\end{equation}

In the $t^{th}$ round of self-correction, the relationship between accuracy before and after self-correction and the two metrics above is given by (with derivation details shown in Appendix \ref{app:math_derivation}):
\begin{equation}
\label{equ:recurrence}
\begin{aligned}
Acc_{t} = Acc_{t-1}CL_{t-1} + (1-Acc_{t-1})CS_{t-1}
\end{aligned}
\end{equation}

Assuming that \textit{CL} and \textit{CS} reflect the inherent confidence and critique capabilities of LLMs, so we treat these metrics as constants independent of the round number $t$, which is empirically validated in Appendix \ref{app:CL_CS_valiadation}, and this yields:

\begin{equation}
\begin{aligned}
\label{equ:recursion_equation1}
Acc_{t} = Acc_{t-1}*CL + (1-Acc_{t-1})*CS
\end{aligned}
\end{equation}

Noticing that Equation \ref{equ:recursion_equation1} and Equation \ref{equ:recursion_equation} are essentially the same recurrence relation, we can similarly derive that:
\begin{equation}
\begin{aligned}
\label{equ:final_equation}
Acc_t &= Upp - \alpha^t( Upp - Acc_0)
\end{aligned}
\end{equation}

where $Upp = \frac{CS}{1-CL+CS}, \alpha=CL-CS$. Empirically we have $0<\alpha<1$, and as $t \rightarrow + \infty$, $Acc_t \rightarrow Upp$.
This equation describes how accuracy changes in multi-round self-correction and provides a theoretical performance upper bound, serving as the core part of our theory.

\subsection{Corollaries}
\label{subsec:corollaries}
Based on our theory, three corollaries can be further derived: 
(1) after infinite rounds of self-correction, the final accuracy converges to the upper bound $Upp$, which is solely determined by $CL$ and $CS$ and is independent of the initial accuracy $Acc_0$; 
(2) the speed of convergence depends $\alpha=CL-CS$, and accuracy converge faster when $\alpha$ is lower; 
(3) in particular, under the ideal condition with an oracle verifier ($CL=1$), the accuracy follows $Acc_t = 1 - (1-CS)^t( 1 - Acc_0)$, ultimately converging to 100\%. 
The derivation details are shown in Appendix \ref{sec:corollaries}.

\begin{figure*}[!htb]
    \centering
    \includegraphics[width=0.95\textwidth]{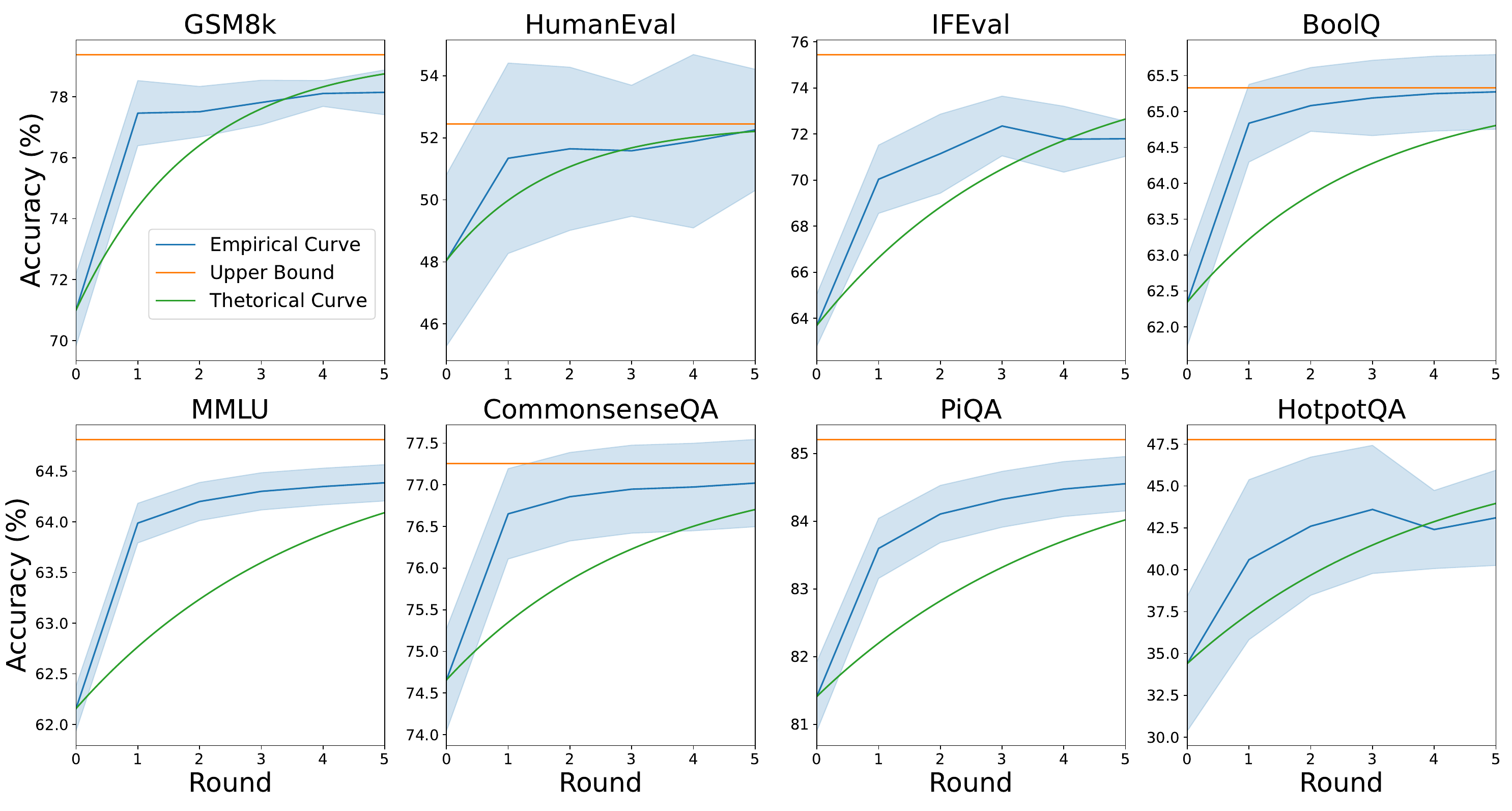}
    \caption{Experimental verification of our theory on Llama3-8B-Instruct. The empirical curve in mult-round self-correction, theoretical curve, and upper bound predicted by our theory are depicted in \textcolor{blue}{blue}, \textcolor{green}{green}, and \textcolor{orange}{orange} respectively. The theoretical curve fits the empirical curve well and accuracy approaches but never surpasses the upper bound.}
    \label{fig:fit}
\end{figure*}

\section{Experiments}

\label{sec:experiment}

\subsection{Experimental Setup}

\paragraph{Models}
Experiments are conducted on both open-source and closed-source models. For the closed-source models, we assess Qwen-Max \citep{qwen}, GPT-3.5 Turbo, and GPT-4 Turbo \citep{gpt4} by API calls. For the open-source models, we evaluate Llama3-8B \citep{llama3}, Qwen2.5-7B \citep{yang2024qwen2}, DeepSeek-LLM-7B \citep{deepseek-llm}, Mistral-7B-v3 \citep{mistral}, and GLM4-9B \citep{glm2024chatglm}, and parameters of these models are publicly available on HuggingFace\footnote{\url{https://huggingface.co/}}.
For each open-source model (< 10B), we run the experiments on a single Nvidia A100 80G GPU, and utilize vllm \footnote{\url{https://github.com/vllm-project/vllm}} to accelerate generation.
Similar to \citet{yang2024decomposition,zhang2024understandingdarkllmsintrinsic}, we adopt "reask" prompt strategy to encourage models to self-correct (i.e. asking the question again).

\paragraph{Dataset}
We conduct experiments on both classification and generation tasks, including GSM8k \citep{cobbe2021gsm8k}, Humaneval \citep{chen2021codex}, IFEval \citep{zhou2023instruction}, MMLU \citep{hendrycks2021measuring}, BoolQ \citep{clark-etal-2019-boolq}, CommonsenseQA \citep{talmor-etal-2019-commonsenseqa}, PiQA \citep{bisk2019piqareasoningphysicalcommonsense}, and HotpotQA \citep{yang-etal-2018-hotpotqa}.

\subsection{Main Results}

To validate our theory, we compare the empirical accuracy change curve with the theoretical curve predicted by our theory by visualizing them in the same figure and checking the alignment.
The empirical curve is acquired from a 5-round self-correction process across multiple models and datasets, during which we track accuracy and variance changes. 
To enhance the numerical stability of experimental results, we sample five responses independently for each question and use the average accuracy for analysis.
For the theoretical curve, we compute three key parameters with a single self-correction: initial accuracy ($Acc_0$), confidence level ($CL$), and critique score ($CS$). Using these values and Equation \ref{equ:final_equation}, we generate the theoretical curve and its upper bound. 
Since the calculation of $CL$ and $CS$ relies on probability, we utilize the probability estimation methods provided by \citet{yang2024decomposition}, and more details are shown in the Appendix \ref{app:probability_estimation}.

The experimental results of Llama3-8B-Instruct are presented in Figure \ref{fig:fit}, 
with more results of other models provided in Appendix \ref{app:experiment_results}. 
The results demonstrate that the theoretical curve closely aligns with the empirical curve across various datasets, suggesting that the proposed theory effectively models and explains the variations in accuracy during self-correction.
Furthermore, the upper bound derived from the theory holds practical relevance, as the accuracy curve consistently approaches but does not exceed it, further validating the effectiveness of our theory.

\subsection{Verification of Corollaries}
We also provide experimental verification of 3 corollaries, which can also serve as further validation of our theory:
(1) we systematically manipulate the initial accuracy to various target values and observe its impact on the final accuracy, finding that the final accuracy consistently converges to the same value (\S \ref{subsec:corollary1});
(2) we compare the convergence rates of models with distinct $\alpha$ values, finding that models with lower $\alpha$ converge noticeably faster (\S \ref{subsec:corollary2})
(3) we equip models with an oracle verifier (CL=1) and observe model performance, finding that model performance boosts fast and finally converges to 100\% (\S \ref{subsec:corollary3}).
Detailed experiment setups and results are provided in Appendix \ref{sec:corollaries}.

\section{Related Work}
\paragraph{Inference Scaling}

Model performance can be improved by introducing more computational cost at test time, and this inference scaling \citep{snell2025scaling,hoffmann2022training} can be achieved via Chain-Of-Thought \citet{wei2022chain}, repeated sampling \citet{wu2024scaling}, Monte Carlo Tree Search\citet{zhang2023planning,liu2024dont}, and multi-round self-correction \citet{liu2024intrinsicselfcorrectioncapabilityllms,xi-etal-2023-self,zhang2024accessinggpt4levelmathematical}.
Our work provides a theoretical framework to understand why and how inference scaling works.

\paragraph{LLM Self-Correction}
LLMs can correct their self-generated answers, and this capability \citep{kamoi-etal-2024-llms,pan-etal-2024-automatically,yang2024decomposition} can be enhanced through external feedback \citep{jiang2023selfevolve}, better prompting strategies \citep{li2024confidence,wu-etal-2024-large}, reinforcement learning \citep{kumar2024training} and iterative self-correction \citep{qu2024recursive,madaan2024self}.
Different from previous works, we propose a theory to explain and model the accuracy curve for self-correction.

\section{Conclusion}

We propose a probabilistic theory to model and explain how accuracy evolves in multi-round self-correction along with 3 corollaries.
Extensive experiments validate the theory by showing the alignment between our theoretical curves and empirical curves, and empirical verification of 3 corollaries also futher supports the theory.
Our theory provides theoretical support and a better understanding of LLM self-correction, thus paving the way for further explorations.

\section*{Limitations}

The calculation of our theoretical curve relies on probability estimation, which necessitates repeated sampling for the same question, 
and the simulation of multi-round self-correction (i.e. actual curve) also generates multiple answers for the same question. These can be more computationally expensive than traditional experiments where only one answer is generated for a question.
We only experimentally validate our theory on 8 models and 8 datasets in the intrinsic self-correction setting, leaving more verification experiments on more datasets (e.g. multi-step reasoning tasks) and setting (e.g. external self-correction) for future work. 

Though our theoretical curve can fit the actual curve to some extent, what happens in self-correction and how accuracy changes can be much more complex than our theory. 
Our theory can only describe how accuracy changes in multi-round self-correction, but how performance improves in other inference scaling settings (e.g. long COT, MCTS) is still unknown, and we leave it to future work.

\section*{Ethical Considerations}

The data we utilized are open for research, and evaluated LLMs are all publicly available by either parameters or API calls. 
Therefore, we do not anticipate any ethical concerns in our research.

\section*{Acknowledgments}
We sincerely thank all anonymous reviewers for their valuable feedback.
This paper is sponsored by State Key Laboratory of Multimedia Information Processing Open Fund.

\bibliography{custom}

\clearpage
\appendix
\section*{Appendix}
\label{sec:appendix}
\section{Mathematical Notations}
\label{app:notations}
This section shows all of the mathematical notations used in our theory. If you forget the meaning of any notation, please refer to Table \ref{tab:notations}. 
We leverage $\ \hat{} \  $ to symbolize estimates (e.g. $\hat{P}(a_i)$ represents the estimate of the true value $P(a_i)$ ). 

\begin{table*}[!tb]
\centering
\footnotesize
\resizebox{\textwidth}{!}{
\begin{tabular}{l|p{0.65\linewidth}}
\toprule
\textbf{Notations} & \textbf{Meanings} \\
\midrule
$Q$ & a dataset with $n$ questions\\
\cmidrule{1-2}
$q_i$ & the $i^{th}$ question in $Q$\\
\cmidrule{1-2}
$a_{i,t}$ & the answer to question $q_i$ generated in the $t^{th}$ round of self-correction \\
\cmidrule{1-2}
$P(a_{i,t})$ & the probability of generating a correct answer for question $q_i$ through a single temperature-based sampling in the $t^{th}$ round of self-correction \\
\cmidrule{1-2}
$P(a_{i,t}|a_{i,t-1})$ & the conditional probability of $a_{i,t}$ is correct given $a_{i,t-1}$ is correct \\
\cmidrule{1-2}
$P(a_{i,t}|\neg a_{i,t-1})$ & the conditional probability of $a_{i,t}$ is correct given $a_{i,t-1}$ is incorrect \\
\cmidrule{1-2}
$P_i^{con}$ & model confidence in question $q_i$: for any $t \in N+$, we have $P(a_{i,t}|a_{i,t-1})=P_i^{con}$ \\
\cmidrule{1-2}
$P_i^{cri}$ & critique capability in question $q_i$: for any $t \in N+$, we have $P(a_{i,t}|\neg a_{i,t-1})=P_i^{cri}$, \\
\cmidrule{1-2}
$P_i^{upp}$ & the upper bound of $P(a_{i,t})$, and we have $P_i^{upp} = \frac{P_i^{cri}}{1- P_i^{con} + P_i^{cri}}$ \\
\cmidrule{1-2}
$\alpha_i$ & the convergence rate of $P(a_{i,t})$, and we have $\alpha_i = P_i^{con} - P_i^{cri}$ \\
\cmidrule{1-2}
$Acc_0$ & the initial accuracy \\
\cmidrule{1-2}
$Acc_t$ & accuracy after the $t^{th}$ round of self-correction \\
\cmidrule{1-2}
$CL$ & the conditional probability of getting a correct answer after self-correction, given the answer before self-correction is correct. (defined in Equation \ref{equ:CL}) \\
\cmidrule{1-2}
$CS$ & the conditional probability of getting a correct answer after self-correction, given the answer before self-correction is incorrect. (defined in Equation \ref{equ:CS}) \\
\cmidrule{1-2}
$Upp$ & the upper bound of $Acc_t$, and we have $Upp = \frac{CS}{1- CL + CS}$ \\
\cmidrule{1-2}
$\alpha$ & the convergence rate of $Acc_t$, and we have $\alpha = CL - CS$ \\

\bottomrule
\end{tabular}
}
\caption{Mathematical notations and their meanings.}
\label{tab:notations}
\end{table*}

\section{Mathematical Derivations}
\label{app:math_derivation}

\subsection{Derivation of Equation \ref{equ:final_question_level}}

First, we have the following equation:

\begin{equation}
\begin{aligned}
P(a_{i,{t}}) &= P(a_{i,t-1})P_i^{con} + [1-P(a_{i,t-1})]P_i^{cri} \\
 &= (P_i^{con}-P_i^{cri})P(a_{i,t-1}) + P_i^{cri} 
\end{aligned}
\end{equation}

By subtracting $\frac{P_i^{cri}}{1-P_i^{con}+P_i^{cri}}$ from both sides of the above equation, we have:
\begin{equation}
\begin{aligned}
&P(a_{i,{t}}) - \frac{P_i^{cri}}{1-P_i^{con}+P_i^{cri}} \\
&=(P_i^{con}-P_i^{cri})(P(a_{i,t-1})- \frac{P_i^{cri}}{1-P_i^{con}+P_i^{cri}})
\end{aligned}
\end{equation}

It is evident that $P(a_{i,t}) - P_i^{upp}$ forms a geometric progression with a common ratio of $\alpha_i$, where $P_i^{upp} = \frac{P_i^{cri}}{1-P_i^{con}+P_i^{cri}}$ and $\alpha_i = P_i^{con}-P_i^{cri}$. By applying the general term formula of a geometric sequence, we obtain: $P(a_{i,t}) - P_i^{upp} = \alpha_i^t(P(a_{i,0})- P_i^{upp})$.

After $k$ rounds of self-correction, the probability of the model correctly answering question $q_i$ is expressed as:
\begin{equation}
\begin{aligned}
P(a_{i,{t}}) &= P_i^{upp} - \alpha_i^t( P_i^{upp} -P(a_{i,0}))
\end{aligned}
\end{equation}

\subsection{Derivation of Equation \ref{equ:recurrence}}
The detailed derivation of Equation \ref{equ:recurrence} is show as follows:

\begin{equation*}
\label{equ:weighted_sum_derivation}
\begin{aligned}
& Acc_t \\
&= \frac{\sum_{i=1}^{n}P(a_{i,t})}{n} \\ 
&= \frac{\sum_{i=1}^{n}P(a_{i,t}|a_{i,t-1})P(a_{i,t-1})}{n} \\
& +\frac{P(a_{i,t}|\neg a_{i,t-1})P(\neg a_{i,t-1})}{n} \\
&= \frac{\sum_{i=1}^{n}P(a_{i,t-1})}{n}\frac{\sum_{i=1}^{n}P(a_{i,t-1})P(a_{i,t}|a_{i,t-1})}{\sum_{i=1}^{n}P(a_{i,t-1})} \\
& + \frac{\sum_{i=1}^{n}[1-P(a_{i,t-1})]}{n} \\
& *\frac{\sum_{i=1}^{n}P(\neg a_{i,t-1})P(a_{i,t}|\neg a_{i,t-1})}{\sum_{i=1}^{n}[1-P(a_{i,t-1})]} \\
&=Acc_{t-1}*CL_{t-1}+(1-Acc_{t-1})*CS_{t-1}
\end{aligned}
\end{equation*}

\section{Validation of Stability of CL and CS}
\label{app:CL_CS_valiadation}

\begin{figure*}[!htb]
    \centering
    \includegraphics[width=0.95\textwidth]{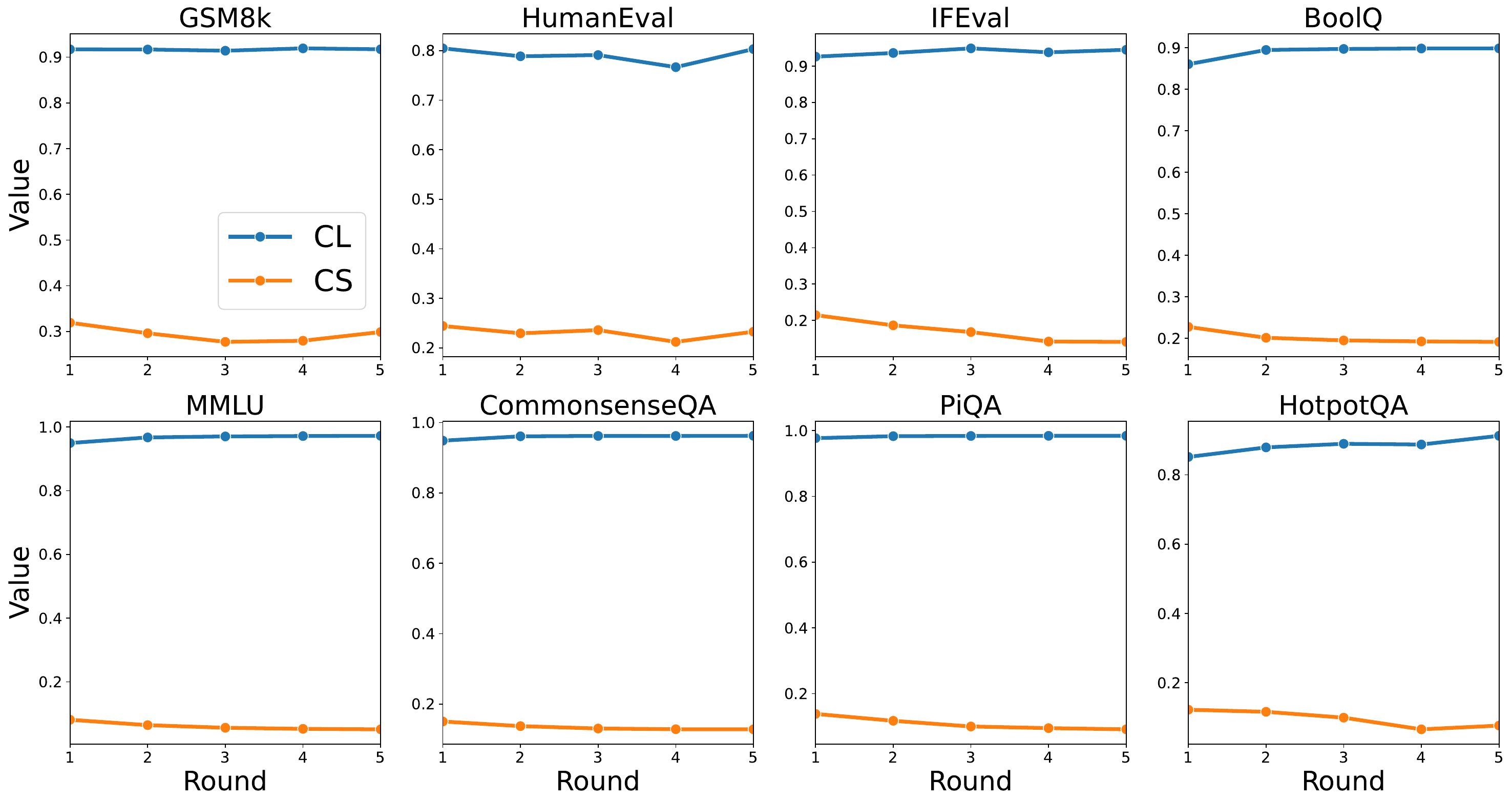}
    \caption{CL of CS values of Llama3-8B-Instruct in different rounds of self-correction.}
    \label{fig:stable}
\end{figure*}

The values of CL and CS can be influenced by the model, dataset, and prompts \citep{yang2024decomposition}. 
We investigate how CL and CS values change with the round of self-correction increases for a given dataset, model, and prompt strategy.
As the results of Llama3-8B-Instruct shown in Figure \ref{fig:stable}, 
CL and CS values remain nearly constant across multiple rounds of self-correction.

\section{Probability Estimation}
\label{app:probability_estimation}

The metrics $CL$ and $CS$ discussed in \S \ref{sec:theory} are derived from a probabilistic perspective and the calculation depends on three key probability values for each question \( q_i \): \( P(a_{i,t}) \), \( P(a_{i,t+1}|a_{i,t}) \), and \( P(a_{i,t+1}|\neg a_{i,t}) \). However, these probabilities are not directly observable. Therefore, we employ statistical methods proposed by \citet{yang2024decomposition} to estimate these probabilities as \( \hat{P}(a_{i,t}) \), \( \hat{P}(a_{i,t+1}|a_{i,t}) \), and \( \hat{P}(a_{i,t+1}|\neg a_{i,t}) \) for metric computation. Natural Language Processing (NLP) tasks are generally divided into classification and generation tasks, and we will separately discuss the probability estimation methods applicable to each type of task.

\paragraph{Probability Estimation for Classification Tasks.} In a \( K \)-class classification task, let the set of all candidate labels be denoted by \( L = \{l_0, l_1, \ldots, l_{K-1}\} \) (e.g., the candidate set for MMLU is \( \{A, B, C, D\} \)). A question \( q_i \) is input into the model, which outputs a predicted label. During next-token prediction, the model generates a logit vector \( (o_0, o_1, \ldots, o_{|V|-1}) \), where each element corresponds to a token in the vocabulary \( V \), whose size is \(|V|\). The logits are then passed through a softmax function to compute the probability distribution for the next token across the entire vocabulary.
For classification tasks, we focus only on probabilities over the candidate label set \( L \), not the whole vocabulary \( V \). Thus, we discard most logits, retaining only those corresponding to candidate labels, producing a reduced logit vector \( (o^{'}_0, o^{'}_1, \ldots, o^{'}_{K-1}) \). After applying the softmax function, the model predicts the probabilities for each label \( P(l_0), P(l_1), \ldots, P(l_{K-1}) \). 

(1) Assuming without loss of generality that the correct label is \( l_0 \), then \( \hat{P}(a_{i,t}) = P(l_0) \).

(2) By feeding the correct answer \( l_0 \) back into the model for self-correction, it outputs a probability distribution over candidate labels, denoted as \( P(l_0|l_0), P(l_1|l_0), \ldots, P(l_{K-1}|l_0) \), leading to \( \hat{P}(a_{i,t+1}|a_{i,t}) = P(l_0|l_0) \).

(3) The computation of \( \hat{P}(a_{i,t+1}|\neg a_{i,t}) \) is more complex. For each incorrect label \( l_j \) (\( j \neq 0 \)), we input it to the model, allowing for self-correction, yielding the probability of correcting to the correct label \( P(l_0|l_j) \). Using the law of total probability, we have \( \hat{P}(a_{i,t+1}|\neg a_{i,t}) = \sum_{j=1}^{K-1} P(l_0|l_j)P(l_j) \).

\paragraph{Probability Estimation for Generation Tasks.} We utilize multiple sampling to estimate probabilities by observing the frequency of correct and incorrect answers. Given a question \( q_i \), we input it to the model to obtain an initial answer, which the model then attempts to self-correct to produce a refined answer. This process is independently repeated \( M \) times, and each pair of initial and refine answers is evaluated for correctness, yielding a sequence of results \((a_{i,t}^0, a_{i,t+1}^0), (a_{i,t}^1, a_{i,t+1}^1), \ldots, (a_{i,t}^{M-1}, a_{i,t+1}^{M-1})\), where \((a_{i,t}^m, a_{i,t+1}^m)\) denotes the outcome of the \( m^{th} \) repetition. Specifically, \( P(a_{i,t}^m) \) and \( P(a_{i,t+1}^m) \) indicate the correctness of the initial and refined answers, respectively. For a correct initial answer $a_{i,t}^m$, \( P(a_{i,t}^m) = 1 \); otherwise, \( P(a_{i,t}^m) = 0 \). The same logic applies to \( a_{i,t+1}^t \). Using these frequencies, we estimate the probabilities as follows:

(1) \(\hat{P}(a_{i,t}) =  \frac{\sum_{m=0}^{M-1} P(a_{i,t}^m)}{M} \);

(2) \( \hat{P}(a_{i,t+1}|a_{i,t}) = \frac{\sum_{m=0}^{M-1} P(a_{i,t}^m) P(a_{i,t+1}^m)}{\sum_{m=0}^{M-1} P(a_{i,t}^m)}\);

(3) \( \hat{P}(a_{i,t+1}|\neg a_{i,t}) = \frac{\sum_{m=0}^{M-1} (1-P(a_{i,t}^m))P(a_{i,t+1}^m)}{\sum_{m=0}^{M-1} (1-P(a_{i,t}^m))}\).

\section{Corollaries}

Based on the theory in \S \ref{sec:theory}, three corollaries can be further derived: (1). the final converged accuracy is independent of the initial accuracy (\S \ref{subsec:corollary1});
(2). the convergence rate of accuracy increases as $\alpha$ decreases (\S \ref{subsec:corollary2});
(3). a special case of the theory where $CL=1$ (\S \ref{subsec:corollary3}).
We provide both mathematical derivation and experimental verification of these corollaries, which can also serve as further validation of our theory.

\label{sec:corollaries}
\subsection{Corollary 1}
\label{subsec:corollary1}

\begin{tcolorbox}[
    colback=gray!5!white,  
    colframe=gray!50!black, 
    left=1mm, right=1mm, 
    top=0.5mm, bottom=0.5mm, 
    arc=1mm
]
\textit{Corollary 1}: The final converged accuracy is exclusively determined by the confidence and critique capabilities (i.e., $CL$ and $CS$), and remains independent of the initial accuracy $Acc_0$.
\end{tcolorbox}

\begin{figure*}[!tb]
    \centering
    \includegraphics[width=0.95\textwidth]{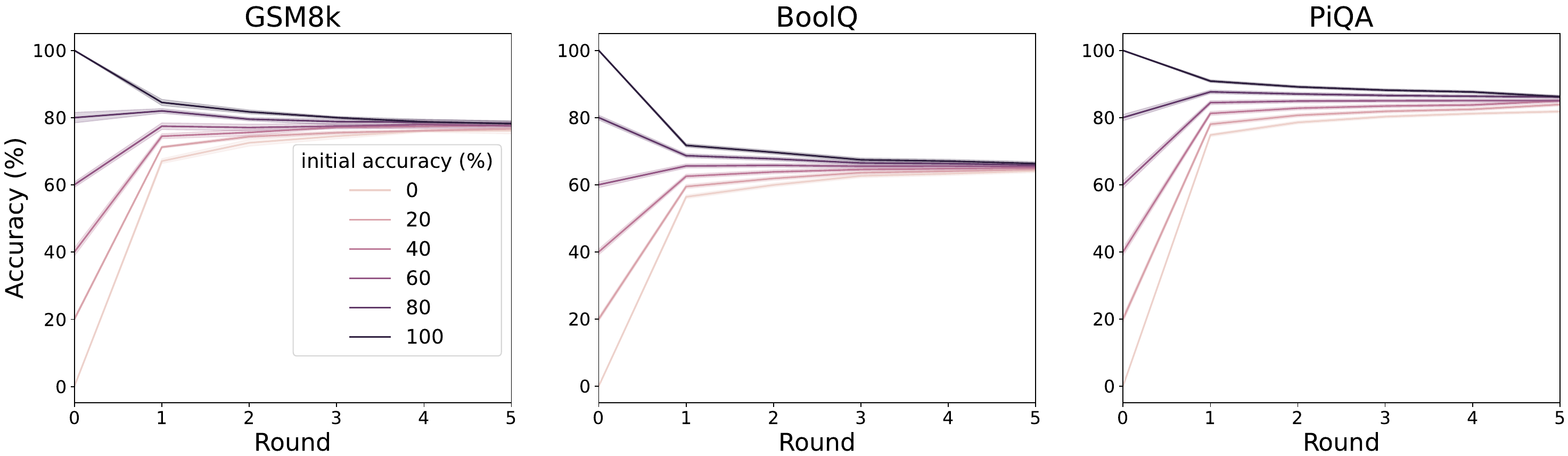}
    \caption{The accuracy convergency results with different initial accuracy $Acc_0$ for Llama3-8B-Instruct: the accuracy consistently converges to the same final value regardless of the initial accuracy.}
    \label{fig:acc_convergence}
\end{figure*}

\paragraph{Derivation of Corollary 1}
Intuitively, when the model is provided with an initial correct or incorrect answer to self-correct, it has a higher probability of reaching the correct answer when the initial answer is correct.  
This implies that $CL > CS$, 
which is also empirically demonstrated by \citet{yang2024decomposition}. 
Given that $CL,CS \in (0,1) $, it follows that $0 < \alpha = CL-CS < 1$.
Based on Equation \ref{equ:final_equation}, as $t \rightarrow + \infty$, $\alpha^t \rightarrow 0$, and thus $Acc_t \rightarrow Upp$.
This indicates after sufficient rounds of self-correction the final accuracy converges to $Upp = \frac{CS}{1-CL+CS}$. Notably, $Upp$ is entirely determined by $CL$ and $CS$ and is independent of the initial accuracy $Acc_0$.

\paragraph{Verification of Corollary 1}
To validate this corollary and investigate whether the initial accuracy influences the final converged accuracy after infinite rounds of self-correction, we systematically manipulate the initial accuracy to various target values and observe its impact on the final accuracy. Unlike the experiments described in §\ref{sec:experiment}, where the initial answer \(a_{i,0}\) is generated by feeding the question \(q_i\) to the model, we directly control the initial accuracy to achieve a desired value \(Acc_{target}\) by carefully setting the initial answers. 
For a \(K\)-class classification task, we assign the initial probability of the correct class to \(Acc_{target}\) and distribute the remaining probability uniformly among the incorrect classes, ensuring that each incorrect class has a probability of \(\frac{1-Acc_{target}}{K-1}\). This guarantees that the initial accuracy \(Acc_0 = Acc_{target}\).  
For generation tasks with \(n\) items in the dataset, we first sample multiple answers for each question \(q_i\) to obtain both correct and incorrect answers. We then randomly select \(\lfloor Acc_{target} \times n \rfloor\) items to use correct answers as initial answers, while assigning incorrect answers to the remaining items, which ensures that the initial accuracy \(Acc_0 \approx Acc_{target}\). In cases where no correct answer is sampled for a question, we use the standard correct answer from the dataset. Conversely, if no incorrect answers are sampled, we truncate a correct answer to create an incorrect one.  
As the results of Llama3-8B-Instruct illustrated in Figure \ref{fig:acc_convergence}, the final accuracy consistently converges to the same value regardless of whether the initial accuracy is set to 0\%, 20\%, 40\%, 60\%, 80\%, or 100\%, which experimentally verifies \textit{Corollary 1}.

\begin{figure*}[!htb]
    \centering
    \includegraphics[width=0.95\textwidth]{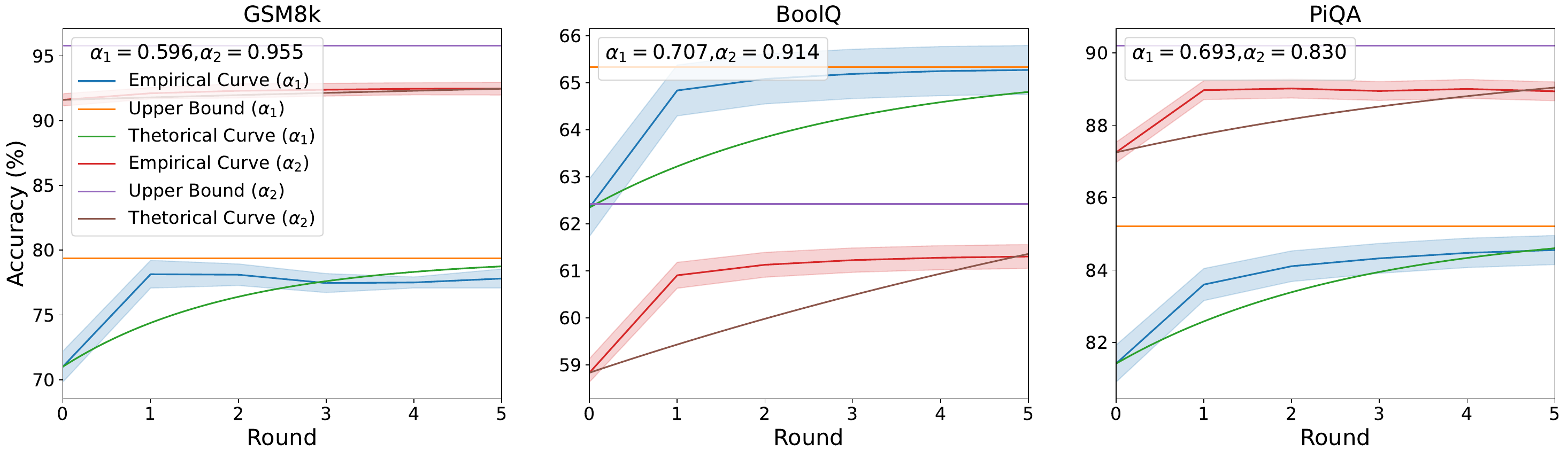}
    \caption{Convergence speed comparison of Llama3-8B-Instruct ($\alpha_1$) and Qwen2.5-7B-Chat ($\alpha_2$): we have $\alpha_1 < \alpha_2$, and Llama3-8B-Instruct ($\alpha_1$) converges noticeably faster and its accuracy gets closer to the upper bound after 5 rounds of self-correction than Qwen2.5-7B-Chat ($\alpha_2$).}
    \label{fig:speed_compare}
\end{figure*}

\begin{figure*}[!tb]
    \centering
    \includegraphics[width=0.95\textwidth]{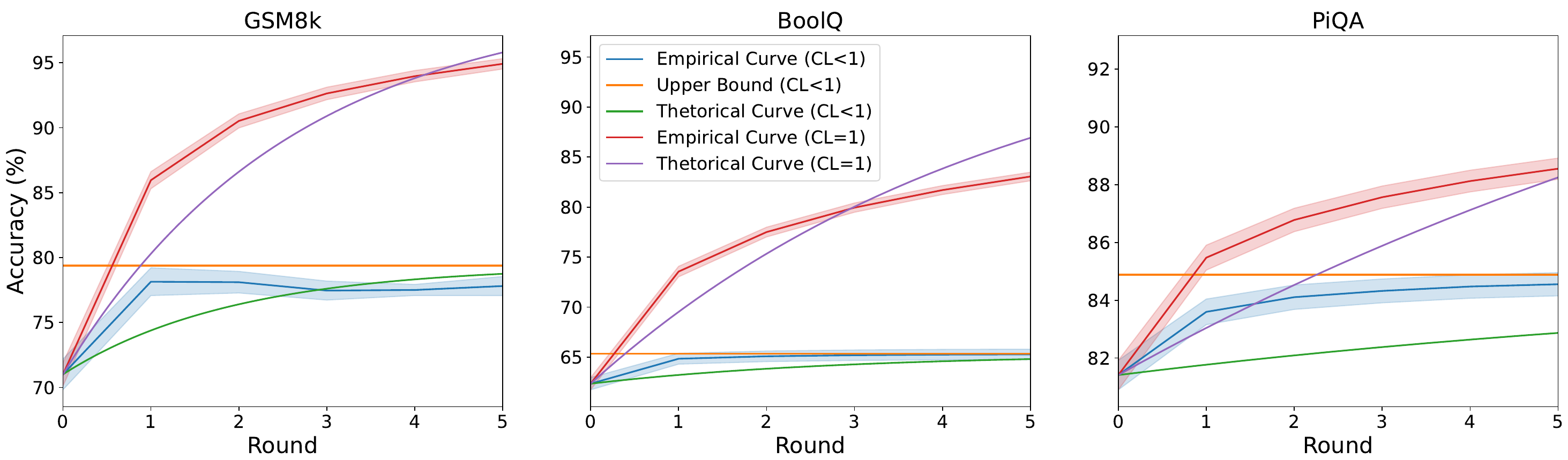}
    \caption{Curves for the special case ($CL=1$) on Llama3-8B-Instruct. The theoretical curve fits the actual curve well when $CL=1$, and exceeds the standard intrinsic self-correction ($CL<1$) by a large margin.}
    \label{fig:special_case}
\end{figure*}

\subsection{Corollary 2}
\label{subsec:corollary2}

\begin{tcolorbox}[
    colback=gray!5!white,  
    colframe=gray!50!black, 
    left=1mm, right=1mm, 
    top=0.5mm, bottom=0.5mm, 
    arc=1mm
]

\textit{Corollary 2}: The convergence rate of accuracy is determined by the parameter $\alpha=CL-CS$. Specifically, a model with a lower value of $\alpha$ exhibits faster convergence in accuracy.
\end{tcolorbox}

\paragraph{Derivation of Corollary 2}
As discussed in \S \ref{subsec:corollary1}, as $t \rightarrow + \infty$, $\alpha^t \rightarrow 0$, and consequently $Acc_t \rightarrow Upp$. The convergence rate of $\alpha^t$ is decided by the value $\alpha$, and the closer the value of $\alpha$ is to $0$, the faster $\alpha^t$ will converge to $0$. 
To better illustrate this difference in convergence speed, consider the following example: when \( \alpha = 0.9 \), \( \alpha^{10} \approx 0.35 \); whereas when \( \alpha = 0.2 \), \( \alpha^{10} \approx 10^{-7} \).

\paragraph{Verification of Corollary 2}
To validate this corollary, we compare the convergence rates of models with distinct $\alpha$ values. 
Given the difficulty in discerning convergence speed differences between models with similar $\alpha$ values, we select two models with significantly differing $\alpha$ values for comparison. 
As experimentally demonstrated in \S \ref{sec:experiment}, the Llama3-8B-Instruct model exhibits a lower $\alpha$ value, while the Qwen2.5-7B-Chat model has a higher $\alpha$ value, so we choose these two models for comparison and analysis. 
The experimental results are shown in Figure \ref{fig:speed_compare}, with more results provided in Appendix \ref{app:experiment_results}.
Llama3-8B-Instruct (lower $\alpha$) converges noticeably faster and its accuracy gets closer to the upper bound after 5 rounds of self-correction than Qwen2.5-7B-Chat (higher $\alpha$), which experimentally verifies \textit{Corollary 2}.

\subsection{Corollary 3}
\label{subsec:corollary3}

\begin{tcolorbox}[
    colback=gray!5!white,  
    colframe=gray!50!black, 
    left=1mm, right=1mm, 
    top=0.5mm, bottom=0.5mm, 
    arc=1mm
]
\textit{Corollary 3}: A special case where CL=1, we have $Acc_t = 1 - (1-CS)^t( 1 - Acc_0)$, and $Acc_t \rightarrow 1$ as $t \rightarrow + \infty$.
\end{tcolorbox}

\paragraph{Derivation of Corollary 3}
For intrinsic self-correction, LLMs need to independently evaluate the correctness of their generated answers \citep{zhang-etal-2024-small}, and errors in this process are almost inevitable \citep{stechly2023gpt,tyen-etal-2024-llms}.
In cases where LLMs incorrectly identify a correct initial answer as erroneous and subsequently generate an incorrect answer after self-correction (\Checkmark $\rightarrow$ \XSolidBrush), we have $CL<1$ instead of $CL=1$.
In contrast, external self-correction helps LLMs determine the correctness of their answers through external feedback, leading to a higher $CL$. 
For instance, \citet{zhang2023planning,kim2023language} employ an oracle verifier to evaluate answer correctness, while \citet{brown2024large} investigate inference scaling laws under the best-of-\textit{n} metric, 
which can be considered as a special case in our theory when $CL=1$. Specifically, when $CL=1$, we have $Upp = \frac{CS}{1-CL+CS}=1, \alpha=1-CS$, yielding:
\begin{equation}
\begin{aligned}
Acc_t &= 1 - (1-CS)^t( 1 - Acc_0)
\end{aligned}
\end{equation}
As $t \rightarrow + \infty$, $\alpha^t \rightarrow 0$, and thus $Acc_t \rightarrow 1$, which aligns with the idea proposed in \citet{brown2024large} that with sufficient times of sampling, the correct answer will always be encountered.

\paragraph{Verification of Corollary 3}
To validate this corollary, we compare whether the accuracy change curve derived from our theory for the ideal scenario ($CL=1$) aligns with the actual experiment curve. 
To simulate this special case ($CL=1$) and equip the model with an oracle verifier, once a correct answer is generated in generation tasks, we halt subsequent rounds of self-correction and directly treat the following answers as correct. 
For classification tasks, we set the conditional probability of selecting the correct/incorrect answer after self-correction given the answer before self-correction is correct to 1/0 (i.e. setting $P(a_{i,t+1}|a_{i,t})=1, P(a_{i,t+1}|\neg a_{i,t})=0$). 
As the experimental results illustrated in Figure \ref{fig:special_case} and Appendix \ref{app:experiment_results}, we show the experimental curve and theoretical curve for the special case ($CL=1$), along with the curves for standard intrinsic self-correction ($CL<1$) for comparison.
The results demonstrate that the theoretical curve can still align well with the empirical curve in this special case ($CL=1$), which experimentally verifies \textit{Corollary 3}. 
Besides, we also find the accuracy of $CL=1$ is improved by a large margin compared to that of $CL<1$ and can exceed the upper bound of $CL<1$, which shows a promising direction for further optimization of self-correction.

\section{Discussion}
\label{sec:disscussion}
\begin{figure}[!t]
    \centering
    \includegraphics[width=0.48\textwidth]{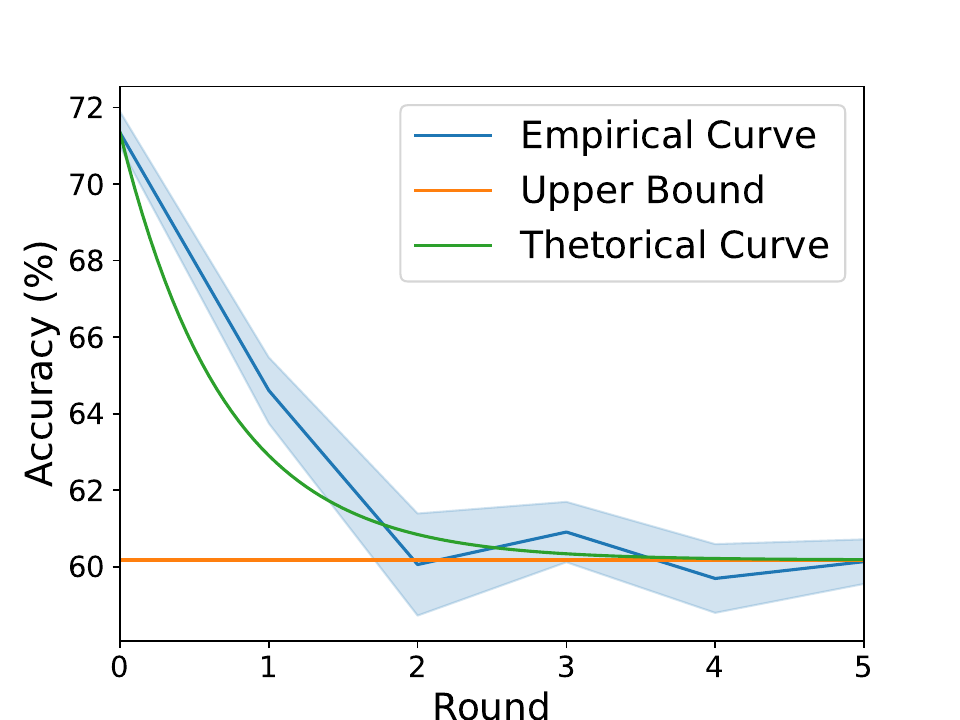}
    \caption{The failure of self-correction of Llama3-8B-Instruct on GSM8k under prompt of "Are you sure?". The accuracy decreases after self-correction and converges to the bound in a descending fashion.}
    \label{fig:fit_fail}
\end{figure}

\paragraph{The Failure of Self-Correction}
Though \citet{madaan2024self,liu2024large} have found LLMs can achieve better performance after self-correction, there is still a debate on the effectiveness of self-correction and \citet{huanglarge,jiang2024self,valmeekam2023can} observe accuracy can even decrease after self-correction with poor prompts.
For instance, \citet{xie-etal-2024-ask,zhang2024understandingdarkllmsintrinsic} find adding "Are you sure?" to the prompt will significantly reduce model confidence, causing it to change correct answers to incorrect ones after self-correction.
Our theory can provide a new perspective to understand how self-correction fails: 
poor prompts can disrupt the balance between the confidence and critique capabilities of LLMs ($CL$ and $CS$), thereby reducing the upper bound ($Upp$) to which the accuracy converges, ultimately resulting in $Upp < Acc_0$, and in this scenario accuracy will decrease after self-correction.
Figure \ref{fig:fit_fail} shows a failure case of Llama3-8B-Instruct on GSM8k under the poor prompt of "Are you sure?", where accuracy converges to the bound in a descending fashion. 
For a given model and test set, different prompts correspond to different $Upp$ values, suggesting that we should choose better prompts to avoid the failure of self-correction. 
A simple approach inspired by our theory could be testing various prompts and selecting the one with the highest $Upp$, and we leave further explorations in avoiding this failure to future work.

\paragraph{How Far Can LLM Self-Correction Go?}
Although previous works \citep{li2024confidence,zhang-etal-2024-self-contrast,wu-etal-2024-large} have utilized and optimized self-correction for better performance, the extent of performance improvements achievable through self-correction under different settings and methods is still not thoroughly explored, and our theory partially fills this gap by providing a theoretical upper bound of accuracy.
Our theory almost announces the death of intrinsic self-correction \citep{xi-etal-2023-self,madaan2024self}, as it demonstrates that intrinsic self-correction cannot surpass the upper bound ($Upp$), which is empirically shown to be not that high in \S \ref{sec:experiment}.
A more promising direction lies in external self-correction \citep{jiang2023selfevolve,chen2024teaching}, as we have discussed in \S \ref{subsec:corollary3} the great performance improvement brought by an oracle verifier (i.e. $CL=1$), and external feedback can be viewed as an approximation of oracle verifier.
Similarly, \citet{kamoi-etal-2024-llms} also discuss this problem and point out future directions for self-correction, and our work provides theoretical support to these discussions.

\section{More Experiment Results}
\label{app:experiment_results}
\begin{figure*}[!htb]
    \centering
    \includegraphics[width=0.95\textwidth]{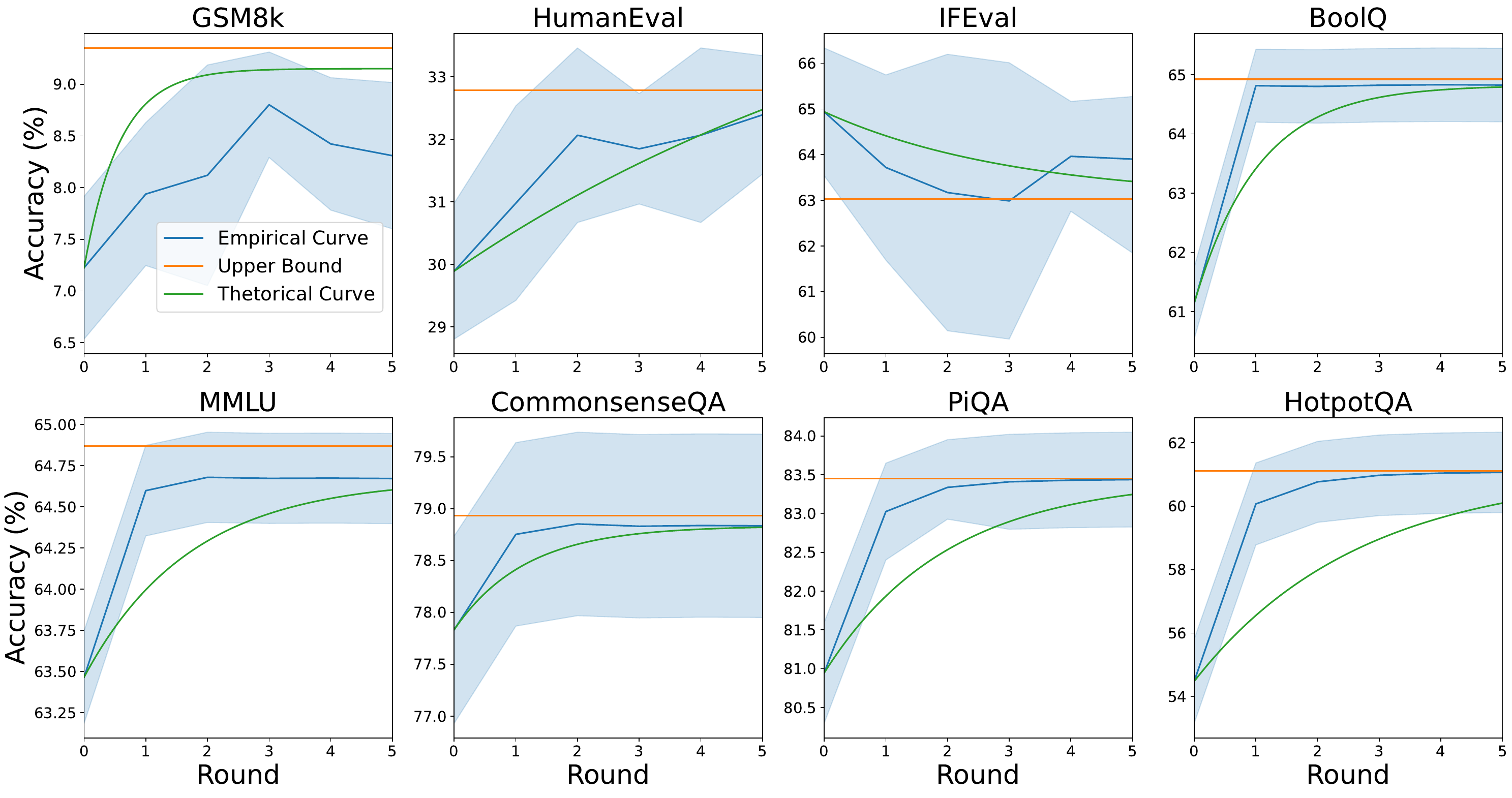}
    \caption{Experimental verification of our theory on BoolQ. The actual curve in mult-round self-correction, theoretical curve, and upper bound predicted by our theory are shown in \textcolor{blue}{blue}, \textcolor{green}{green}, and \textcolor{orange}{orange} respectively. The theoretical curve fits the actual curve well.}
    \label{fig:fit_glm}
\end{figure*}

\begin{figure*}[!htb]
    \centering
    \includegraphics[width=0.95\textwidth]{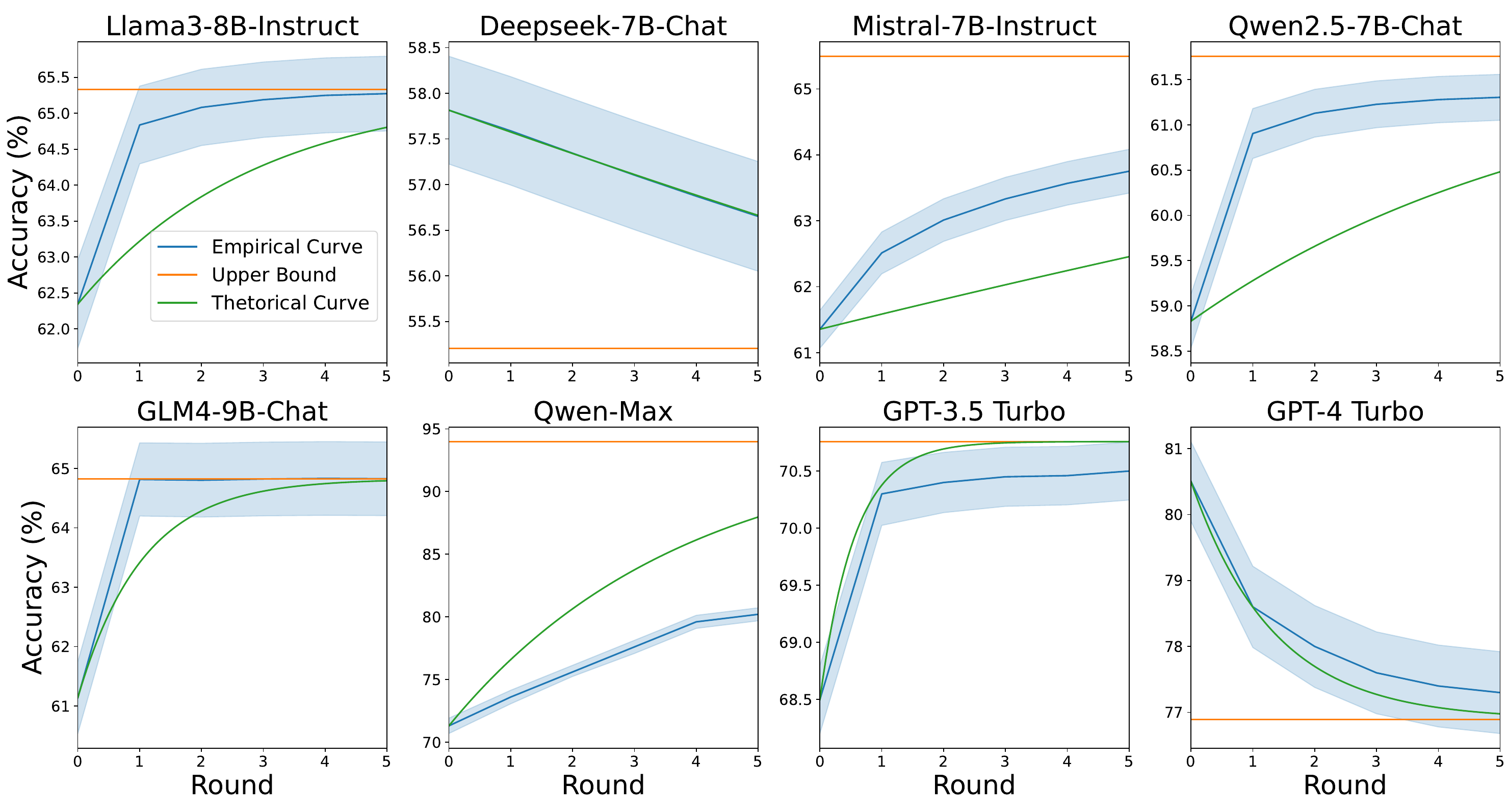}
    \caption{Experimental verification of our theory on GLM4-9B-Chat. The actual curve in mult-round self-correction, theoretical curve, and upper bound predicted by our theory are shown in \textcolor{blue}{blue}, \textcolor{green}{green}, and \textcolor{orange}{orange} respectively. The theoretical curve fits the actual curve well.}
    \label{fig:fit_models}
\end{figure*}

We try to verify on 8 models and 8 datasets in \S \ref{sec:experiment}, but full experiments include $8*8=64$ groups, which is extremely expensive. So we only do a part of them and we believe that is sufficient to validate our theory. 
We show the results of 8 datasets on GLM4-9B-Chat in Figure \ref{fig:fit_glm}, and we also show the results of 8 models on BoolQ in Figure \ref{fig:fit_models}, leaving more validation experiments on other models and datasets to further work.

\begin{figure*}[!htb]
    \centering
    \includegraphics[width=0.95\textwidth]{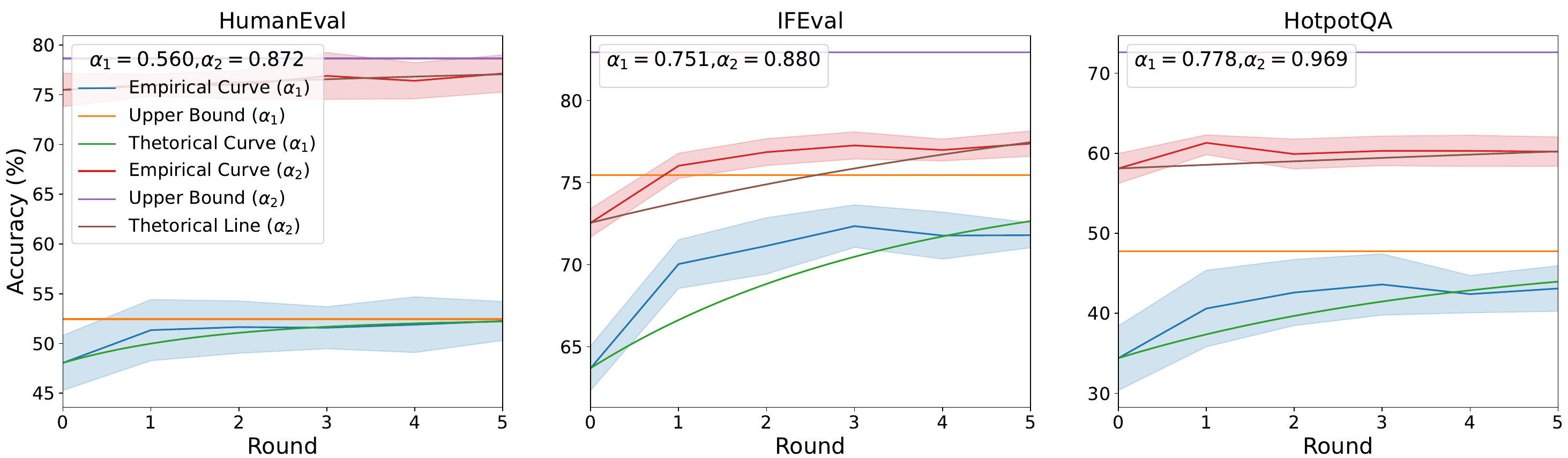}
    \caption{Convergence speed comparison of Llama3-8B-Instruct ($\alpha_1$) and Qwen2.5-7B-Chat ($\alpha_2$): we have $\alpha_1 < \alpha_2$, so Llama3-8B-Instruct ($\alpha_1$) converges noticeably faster and its accuracy gets closer to the upper bound after 5 rounds of self-correction than Qwen2.5-7B-Chat ($\alpha_2$).}
    \label{fig:compare1}
\end{figure*}

\begin{figure*}[!htb]
    \centering
    \includegraphics[width=0.95\textwidth]{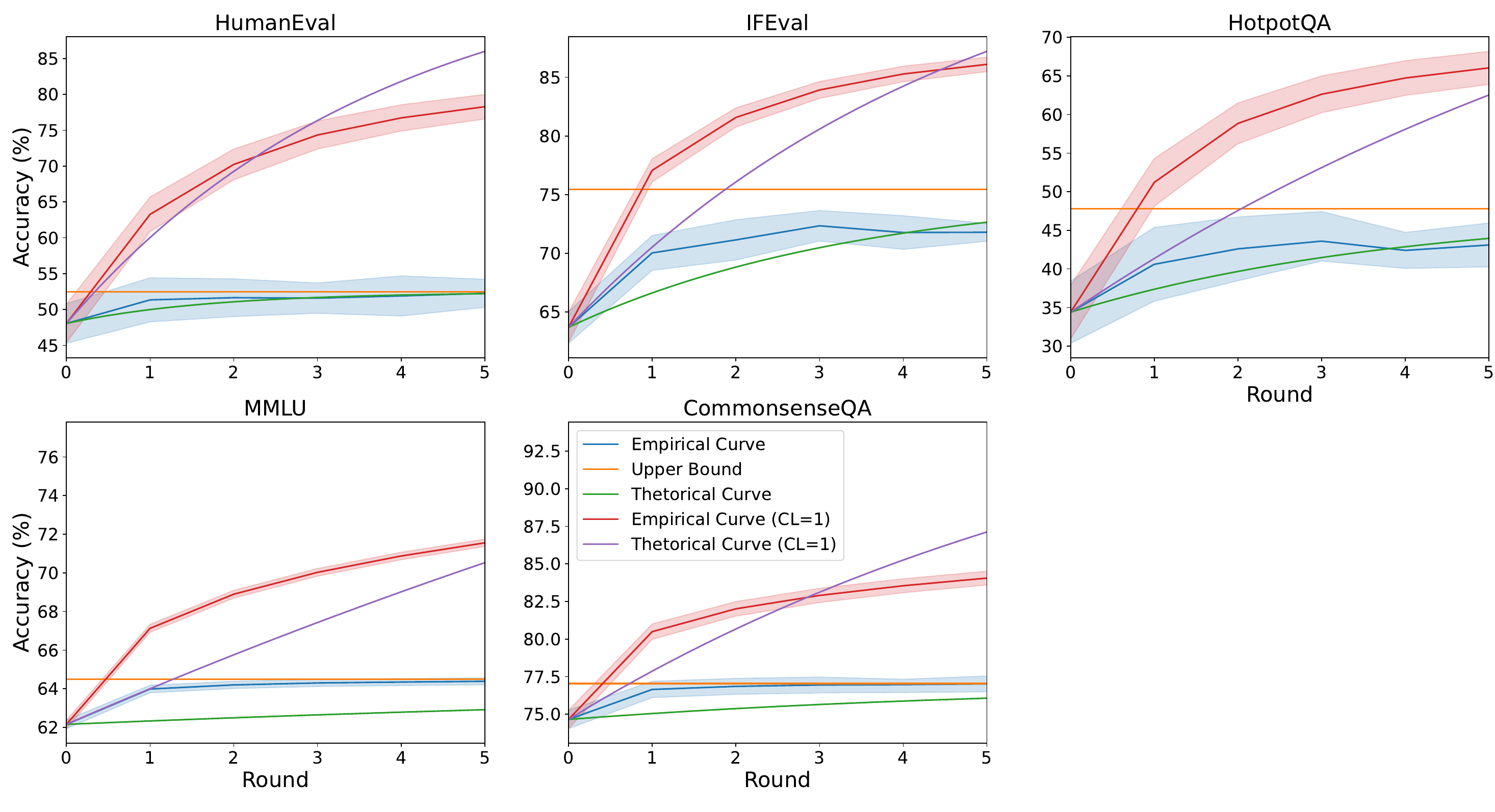}
    \caption{Curves for the special case ($CL=1$) on Llama3-8B-Instruct. The theoretical curve fits the actual curve well when $CL=1$, and exceeds the standard intrinsic self-correction ($CL<1$) by a large margin.}
    \label{fig:special_case1}
\end{figure*}

Except for the main experiments, we also provide more results on the validation of corollaries (\S \ref{sec:corollaries}). More results on convergence rate (\S \ref{subsec:corollary2}) are shown in Figure \ref{fig:compare1}, and more results on a special case where $CL=1$ (\S \ref{subsec:corollary3}) are illustrated in Figure \ref{fig:special_case1}.

\end{document}